\documentclass{article}
\usepackage{acra}
\usepackage{graphicx}
\usepackage{microtype}
\usepackage{array}
\usepackage{xcolor}
\usepackage{enumitem}
\definecolor{hyperblue}{RGB}{36, 116, 242}
\usepackage[colorlinks=true, allcolors=hyperblue, citecolor=.
]{hyperref}
\usepackage{booktabs}
\usepackage{subfigure}
\newcolumntype{L}[1]{>{\raggedright\let\newline\\\arraybackslash}m{#1}}
\newcolumntype{C}[1]{>{\centering\let\newline\\\arraybackslash\hspace{0pt}}m{#1}}
\title{Racing With ROS~2\\A Navigation System for an Autonomous Formula Student Race Car}
\author{Alastair Bradford, Grant van Breda, Tobias Fischer\thanks{The authors acknowledge support by the QUT Centre for Robotics, and T.F. has also been supported from an ARC Laureate Fellowship FL210100156 to Michael Milford, and a grant from Intel Labs via grant RV3.290.Fischer.} \\ Queensland University of Technology\\ 
tobias.fischer@qut.edu.au}

\begin{document}

\maketitle

\begin{abstract}
The advent of autonomous vehicle technologies has significantly impacted various sectors, including motorsport, where Formula Student and Formula: Society of Automotive Engineers introduced autonomous racing classes. These offer new challenges to aspiring engineers, including the team at QUT Motorsport, but also raise the entry barrier due to the complexity of high-speed navigation and control. This paper presents an open-source solution using the Robot Operating System 2, specifically its open-source navigation stack, to address these challenges in autonomous Formula Student race cars. We compare off-the-shelf navigation libraries that this stack comprises of against traditional custom-made programs developed by QUT Motorsport to evaluate their applicability in autonomous racing scenarios and integrate them onto an autonomous race car. Our contributions include quantitative and qualitative comparisons of these packages against traditional navigation solutions, aiming to lower the entry barrier for autonomous racing. This paper also serves as a comprehensive tutorial for teams participating in similar racing disciplines and other autonomous mobile robot applications.
\end{abstract}

\section{Introduction}
Formula Student (FS) and Formula: Society of Automotive Engineers (F-SAE) are international student engineering competitions where students design, manufacture, and race Formula-style vehicles. In 2017, Formula Student Germany introduced an autonomous vehicle racing class~\cite{FSG2017Driverless} to prepare students for challenges in the evolving automotive sector. This initiative led to similar autonomous categories in other competitions, such as the Australasian competition which began in 2021, where vehicles navigate unseen tracks aiming for the fastest overall time across 10 laps~\cite{FSG2023Rules}.

\begin{figure*}[!t]
    \centering
    \includegraphics[width=1\linewidth]{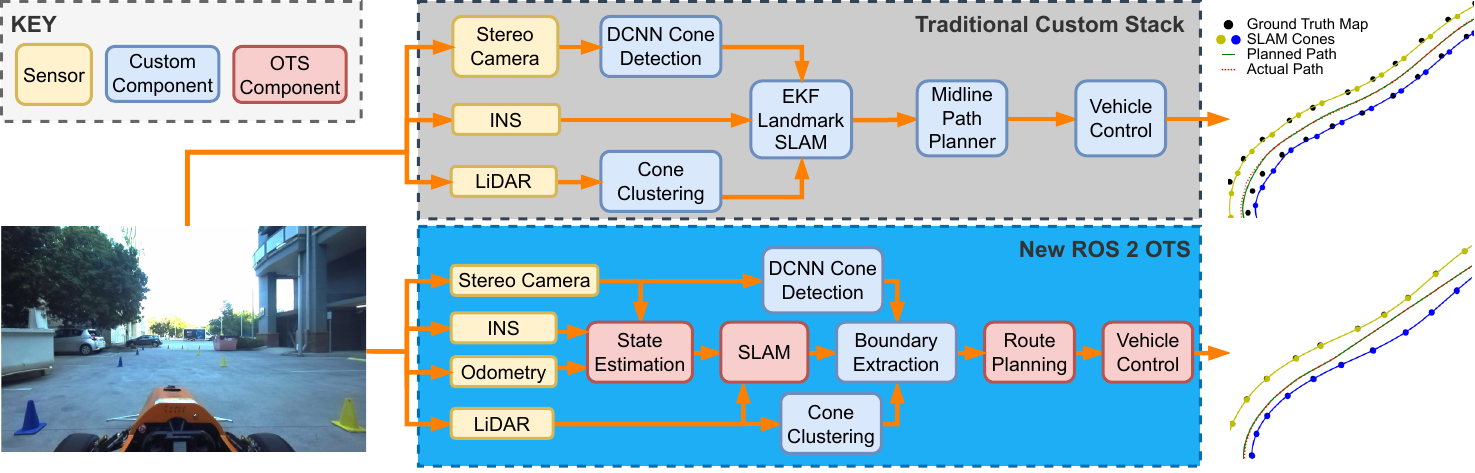}
    \caption{QUT Motorsport's previous custom stack (grey box) compared to our new off-the-shelf ROS~2 Navigation Stack (blue box). Using the ROS~2 Navigation Stack reduces reliance on custom-made software by the team and allows for more complex and mature replacements to be layered together.}
    \label{fig:custom_vs_ros_architecture}
\end{figure*}

Despite these advancements, the barrier to successful autonomous racing remains high, due to the complex nature of high-speed navigation with limited sensor suites and computing power.
The authors, part of the QUT Motorsport team, aim to address these challenges. The team has hitherto seen limited success with autonomous racing solutions, and this paper presents new ways to achieve the team's competition goals. The existing software system on QUT's race car, seen in Figure~\ref{fig:custom_vs_ros_architecture}, consists of custom-developed programs that process sensor data and actuate vehicle movement. These programs are based on foundational university courses and lack the sophistication needed for reliable performance. Substantial time and experience would be needed to elevate these algorithms to a mature and dependable state. %

\begin{figure}[!t] 
\centering
\includegraphics[width=0.9\linewidth]{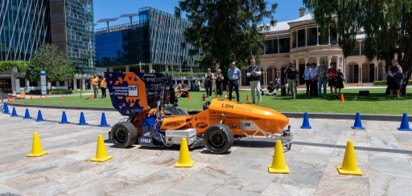}
\caption{QEV-3D performing an autonomous driving demonstration at QUT Gardens Point.}
\label{fig:qev3_demo}
\end{figure}

This paper focuses on leveraging existing off-the-shelf software packages developed by the open-source Robot Operating System (ROS) community. ROS~2 (and its predecessor ROS~1) serves as the de-facto standard for robotic system deployment as it provides a mature middleware to interface between algorithms, machines, and data visualisation~\cite{quigley2009ros,macenski2022ros2_design_architecture,FischerRAM2021}. In addition, ROS ships with a plethora of software programs developed by the open-source community for use in developers’ projects. However, the QUT Motorsport team had not explored any off-the-shelf ROS software for use in autonomous racing prior to this project. In this paper, we focus on replacing our existing algorithms for sensor processing, navigation, and control. 

We hypothesise that using off-the-shelf software leads to higher-performing solutions that can be developed upon more rapidly compared to developing in-house solutions. We also argue that off-the-shelf software improves safety and reliability, which is particularly critical in a high-speed autonomous environment. 

Our contributions are as follows:
\begin{enumerate}[topsep=0pt]
    \item We propose an open-source, replicable ROS~2 Navigation stack for autonomous FS race cars, aiming to lower entry barriers for other teams. This paper serves as a tutorial for similar racing disciplines and autonomous mobile robot applications.
    \item We evaluate the suitability of off-the-shelf ROS~2 navigation packages for autonomous racing and discuss their integration with existing systems.
    \item We provide both qualitative and quantitative comparisons of these packages against previous autonomous navigation solutions, using various accuracy metrics.
\end{enumerate}
We make our code publicly available at: \\
\href{https://github.com/QUT-Motorsport/QUTMS_Nav_Integration}{github.com/QUT-Motorsport/QUTMS\_Nav\_Integration}.

\section{Related Work}

The autonomous navigation landscape has seen significant growth in recent years, particularly with the advent of autonomous mobile robots and autonomous vehicles for road use~\cite{yasuda2020autonomous}. This burgeoning field has led to a multitude of approaches and methodologies aimed at solving the intricate problems of mapping, localisation, route planning, and guidance. This section delves into these key areas, providing an overview of existing solutions and their relevance to autonomous racing.

\subsection{Mapping and Localisation}
Mapping and localisation are foundational elements in the domain of autonomous navigation. Simultaneous Localisation and Mapping (SLAM) algorithms offer a compelling solution by allowing a robot to create a map of its environment while keeping track of its location~\cite{thrun2005probabilistic_robotics,Tsintotas_2022,bresson2017simultaneous,Cadena_2016}. In real-time scenarios, such as racing, SLAM algorithms are crucial for adapting to rapid changes in the environment and for correcting sensor noise, particularly from odometry tracking devices like inertial measurement units (IMU).

One variation of SLAM, known as extended Kalman filter (EKF) landmark SLAM, has been previously employed by the QUT Motorsport team~\cite{suenderhauf2015rss_mobile_sensors_slamm}. This technique involves creating a map of unique object features, such as traffic cones, which are then used to update the robot's pose. The EKF algorithm uses data from an internal inertial navigation system (INS) to predict the vehicle's pose and adjusts it relative to the identified landmarks in the map.

\subsection{Route and Path Planning}
Once a reliable map has been established and the robot's position within it ascertained, the focus shifts to planning a navigable route. In the context of autonomous racing, this involves optimising the route to adhere to the dynamics and constraints of the vehicle~\cite{heilmeier2020minimum_curvature_trajectory_planning_and_control}. Unlike generic path planning algorithms, racing scenarios require specialised algorithms that consider speed, vehicle dynamics, and track conditions.

However, prior implementations by QUT Motorsport have been simplistic, relying on a basic algorithm that calculates a path based on the colour of traffic cones denoting track boundaries. This approach has limitations, particularly when it comes to optimising for speed and manoeuvrability.

\subsection{Guidance}
Guidance is the final piece of the navigation puzzle, translating planned routes into actionable control commands for the robot. Simple algorithms like Pure Pursuit are often employed for this purpose~\cite{elbanhawi2018receding_horizon_pure_pursuit}. Pure Pursuit targets a waypoint ahead of the vehicle and adjusts the vehicle's control settings to reach it. However, this method has limitations, as it does not account for the vehicle's kinematics when pursuing the waypoint.

\subsection{Competitor's Solutions}
A review of existing FS Driverless teams reveals a common infrastructure built on ROS or ROS~2~\cite{alvarez2022software,tian2018autonomous_driving_system_fsd,zeilinger2017fsd}. While these teams employ a variety of sensors, their system architectures often follow a standard perception-planning-control pipeline. Teams in other autonomous racing disciplines, such as F1Tenth and evGrand Prix Autonomous~\cite{cui2020enhanced_safe_and_reliable_autonomous_driving_platform,hall2022occupancy_grid_based_reactive_planner}, also utilise ROS with the same processing pipeline breakdown.

Most of these teams opt for custom-developed software, highlighting a gap in leveraging ROS for a complete autonomous racing solution. 
This survey of related work underscores the importance of developing a comprehensive, ROS-based solution to address the existing gaps in autonomous racing. It is this need that the present paper aims to fulfill.

\section{Preliminaries}
\label{sec:preliminaries}
This section provides an overview of key technologies and frameworks that form the foundation of this paper, particularly focusing on ROS~2 and associated packages for autonomous navigation.
ROS and its successor ROS~2 are middleware solutions designed to facilitate a range of robotic applications, from research to commercial deployments~\cite{macenski2022ros2_design_architecture}. While using ROS as a middleware is a valid use-case on its own, ROS also offers a suite of algorithms developed by the community. Notably, the Nav2 stack—an advanced version of the ROS~1 Navigation stack—bundles various navigation packages cohesively~\cite{macenski2020marathon_2_navigation_system}. This enables simultaneous execution and inter-package communication for map, path, and control data. 

\subsection{Nav2 Overview}
The improvements in Nav2 include managed lifecycle nodes, enhanced composition capabilities, and better networking through Data Distribution Services (DDS). The main advantage of Nav2 over alternative navigation systems was its integration of the BehaviorTree CPP V3 library \cite{BehaviorTreeCPP}. This provided a foundation for robust ROS~2 action services to coordinate navigation tasks from mapping and planning, to control and avoidance.

At its core, Nav2 relies on other ROS~2 tools for robot state estimation and 2D map generation~\cite{macenski2023survey,macenski2020marathon_2_navigation_system}. These include the \verb|robot_localization| and \verb|slam_toolbox| packages that are introduced in the next sections.

\subsection{State Estimation}
\label{sec:prelim_state_estimation}
The \verb|robot_localization| package offers an EKF implementation that allows for the fusion of multiple sensor measurements, such as GPS, IMU, and wheel encoders, to provide a more accurate pose estimate~\cite{moore2016generalized_ekf}, where any one sensor by itself would not provide an accurate state estimate. This package improves upon existing ROS packages by incorporating 3D localisation, expanding sensor inputs, and utilising ROS standard odometry messages for broad software compatibility.

\subsection{Mapping and Localisation}
The \verb|slam_toolbox| package~\cite{macenski2021slam_toolbox}
was designed around the Karto mapping algorithm which can build 2D LiDAR occupancy grid maps while tracking a robot’s pose graph. Here, `map' cells of a defined area would be occupied if LiDAR points registered an object in that area and cells would be free if there were no points that fell there. This limits the scan matching to 2D, which may be adequate in a planar racetrack scenario. The \verb|slam_toolbox| also relies on an estimated robot pose from the \verb|robot_localization| package or similar. The VSLAM ROS packages provide an alternative approach that escape the 2D limitations. A comparative study of VSLAM and the \verb|slam_toolbox| is detailed in~\cite{merzlyakov2021comparison_of_vslam}.

One such technique, RTAB-Map, stood out for its ability to generate dense 3D point clouds with 3D LiDAR scanners, allowing more features to track, which showed improved localisation~\cite{labbe2019rtab_map_lidar_vslam}. However, the risk with creating dense point clouds, potentially hundreds of meters across is memory management and the ability to process data rapidly in real time.

\section{Methodology and Tutorial}
\label{sec:methodology}

This section outlines the methodology for integrating the Nav2 navigation stack into QUT Motorsport's race car. The integration is structured into four phases—state estimation, mapping and localisation, planning, and guidance\footnote{We note that while the Nav2 library compasses route planning, path planning, and dynamic obstacle avoidance with guidance controllers, we have separated these into two distinct phases for testing purposes. Our previous stack had planning and guidance control as separate processes and as found in literature, competing teams structured their navigation systems similarly.}. These phases align with core components discussed in Section~\ref{sec:preliminaries} and are designed to replace or augment existing custom solutions incrementally. The phased approach allowed us to replace one component of the previous custom navigation system at a time, such that we could independently test how each component improves the overall system, as shown in Figure~\ref{fig:custom_vs_ros_architecture}. We also recognise the efforts of other teams that might only want to replace some of their components.

\subsection{State Estimation}
The state estimation stage involved deploying \verb|robot_localization| to estimate the internal odometry of QEV-3D on a competition racetrack. We hypothesised that having a strong robot location estimation would assist and improve the outcome of any chosen mapping algorithm. This was not a replacement for any custom algorithm, rather a feature, which would allow multiple vehicle sensor measurements to be taken into account while estimating state.

The goal of this stage was for the previous navigation system to utilise a more accurate estimate of the vehicle’s state than using single INS odometry estimate. Success criteria for this would see an improvement using off-the-shelf packages across two performance metrics, namely 1) accuracy of the map against a ground truth (RMSE), and 2) accuracy of the estimated trajectory against a ground truth (RMSE).

The integration of \verb|robot_localization| required some adjustment to the existing SLAM as it currently used the changes in INS position to predict the robot state in advance before being updated by external sensor measurements (LiDAR and camera). Using separate odometry filtering meant the mapping algorithm input was now a filtered state estimation, rather than just the GNSS, so this was adjusted accordingly. The sensor measurement update step in the EKF remained the same. 

\subsection{Mapping and Localisation}
Here we leveraged ROS~2 SLAM packages to create a racetrack map and refine QEV-3D's pose estimate with sensors available on the vehicle, thereby replacing QUT Motorsport's previous EKF SLAM algorithm. The success criteria include performance improvements in mapping accuracy and robustness over the same metrics investigated during the state estimation stage.

Specifically, we substituted our current landmark-based SLAM with an occupancy grid mapping approach. The off-the-shelf package utilised for this purpose was the \verb|slam_toolbox|, which accepts laser scans and filtered odometry estimates as inputs. Like in the state estimation stage, the filtered odometry estimates were generated by the \verb`robot_localization` EKF sensor fusion algorithm.

Given that the \verb|slam_toolbox| relies on laser scans rather than cone landmarks for mapping, a separate mechanism was needed to integrate cones into the map. This was a requirement of the competition, where cone detection accuracy was scored \cite{fsaeaRules2023}. Consequently, a custom strategy was developed to ensure compatibility with the existing pipeline (see Section~\ref{sec:custom_interfaces}).

\subsection{Route Planning}
\label{sec:method_planning}
The objective of the route planning phase was to integrate a ROS~2 path planner that generates an optimised route -- a racing line -- around a racetrack, replacing QUT Motorsport's simplistic mid-line planner. The success criteria for this planner focused on two key racing line metrics: 1) cumulative curvature compared to our previous planner, and 2) path distance compared to our previous planner.

This stage introduces advanced planning techniques aimed at route optimisation. The off-the-shelf package utilised here was from Nav2's suite of planning plugins, with configuration options that account for vehicle dynamics, minimal curvature paths, and boundary exclusion.

Nav2's \verb|smac_planner_hybrid| A* global path planner package \cite{macenski2023survey} was chosen due to its use of 2D occupancy grids and granular customisation, particularly for Ackermann steering vehicles. One can specify vehicle dynamics parameters, such as minimum steering radius, and apply penalties to racing lines while the solver iterates through options. In autonomous racing, we require racing lines which push the limits of the track to shorten the distance travelled but maintain minimal curvature so speed is consistently as fast as possible.

\subsection{Guidance and Avoidance}
To conclude the navigation stack, we integrated a ROS~2 guidance controller with adaptive obstacle avoidance serving as an alternative to QUT Motorsport's previous solution. Our success criteria are two key guidance metrics for racing, namely 1) accuracy, as measured as cross-track error against the desired path, and 2) lap-time compared to our previous approach.

The \verb`regulated_pure_pursuit` package from Nav2's controller library~\cite{macenski2023regulated,macenski2023survey} was chosen as the successor to our previous pure pursuit implementation. Although pure pursuit is typically a straightforward and simplistic algorithm, we chose this for our testing over Nav2's more complex guidance controllers such as model predictive path integral control. This was due to the team's limited validation data for vehicle modelling, which lowered our confidence in configuring a more complex controller. Still, \verb`regulated_pure_pursuit` provides many more configuration options for generic vehicle motion than our previous custom developed algorithm.

Additionally, we aimed to handle dynamic collision avoidance while operating within the control loop, which would re-plan the path with the chosen Nav2 planner. This would allow us to account for drift on the map or map error if a traffic cone was hit or knocked out of place slightly during a previous lap. 
We leveraged Nav2's built-in behaviour tree manager to trigger this action at a fixed rate, which involved customising a behaviour tree configuration file to coordinate the planner and controller cohesively.

\subsection{Custom Interfaces}
\label{sec:custom_interfaces}
Despite our best efforts, there were inevitable gaps between the off-the-shelf ROS~2 navigation packages, the existing software infrastructure on QEV-3D, and competition regulation requirements. These were patched with a minimal number of custom ROS~2 packages to ensure complete functionality and compatibility.

The cone placement and landmark map building strategy involved extracting traffic cones using QEV-3D's existing visual sensor processing algorithms and adding them to a list of cone landmarks. With each pose update from the ROS~2 SLAM output, cones in view are searched in the existing landmark list for a `best match'. If a match is found within a specified search radius of 1.5m, then its location is updated based on a moving average; otherwise, a new cone is added.
Additionally, the competition track layout specifies traffic cones may be spaced up to five metres apart (see Section~\ref{sec:experimentalsetup}). As a result, the output of the ROS~2 SLAM package would be an occupancy grid of sparsely occupied cells with gaps between them being large enough for a vehicle to pass through. 

A lack of continual bounds meant ROS~2 occupancy grid planners would attempt to plan through the gaps in the boundary-marking cones. 
Using the landmark map which recorded detected cone locations, continual bounds could be interpolated between cones and a new occupancy grid could then be created. This was published for the Nav2 planner to use rather than the 2D SLAM output occupancy grid.

Nav2's default stack provides behaviour tree customisation in the form of an XML file. This default file provided additional management, recovery, and re-planning  functions within the library that allows for a cohesive end-to-end robotic system. However, this resulted in substantial overhead with processes not required for our simple constrained system. Thus, a simplified custom behaviour tree was configured to manage planning and guidance alone, additionally reducing the number of processes running. %

\section{Experimental Setup}
\label{sec:experimentalsetup}

This section details the experimental framework employed to validate 
our ROS~2 navigation stack's efficiency. The primary objective is to 
assess the performance of the proposed stack in a controlled and 
replicable environment, encompassing both real-world track navigation 
and simulated scenarios. This ensures a comprehensive understanding of 
our testbed.

\subsection{Target Track Specifications}
\begin{figure}[!tb] 
    \centering
    \includegraphics[width=\linewidth]{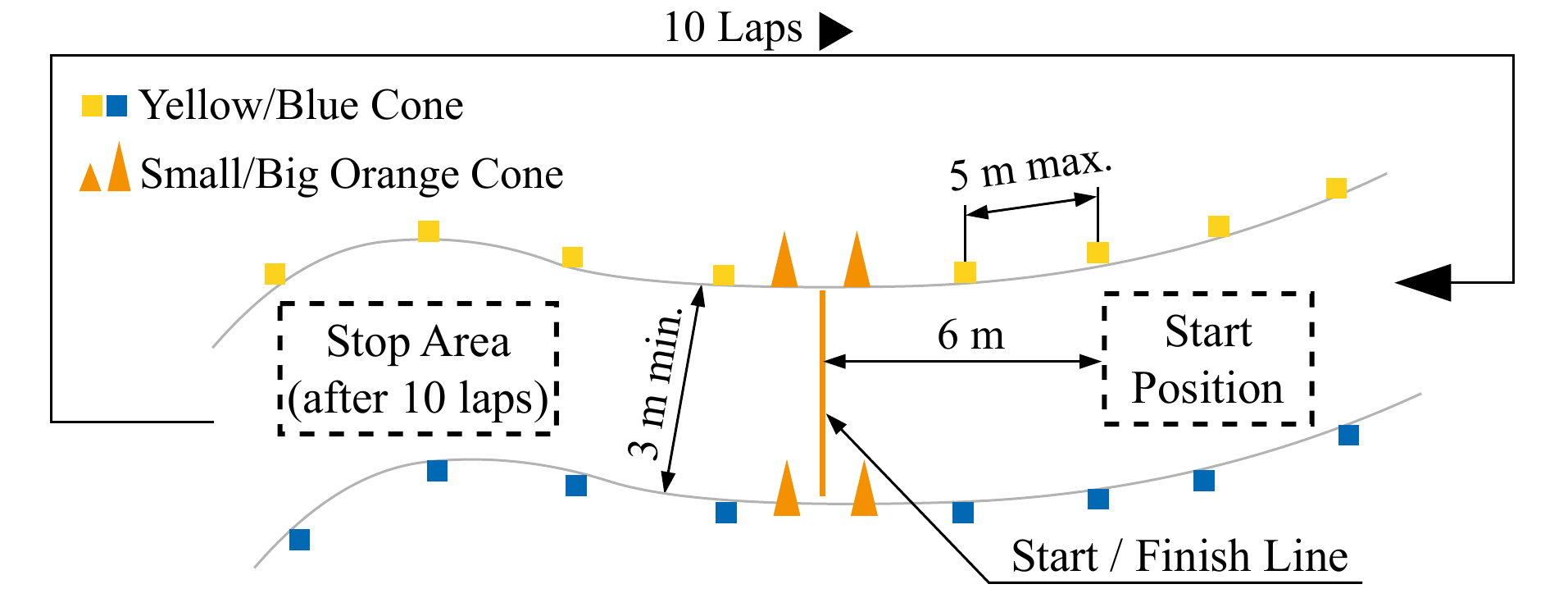}
    \caption{Example track segment adapted from the Formula Student Germany Competition Handbook~\protect\cite{FSG2023Handbook}.}
    \label{fig:track}
\end{figure}

Our research aims to replicate the conditions of a track that mirrors the standard regulations set by the FS competition:

\begin{itemize}
    \item \textbf{Track Boundaries}: Blue cones on the left and yellow cones on the right.
    \item \textbf{Cone Spacing}: Cones spaced at a maximum of 5 meters apart on the same side.
    \item \textbf{Track Width}: Varies between 3 to 6 meters.
    \item \textbf{Start/Finish Line}: Defined by four large orange cones.
\end{itemize}
An example segment of such a track is illustrated in Figure~\ref{fig:track}.

\subsection{Development Platform}
Our choice of development tools and software platforms was guided by both their robustness and relevance to our research objectives:

\begin{enumerate}
	\item \textbf{ROS~2 Humble}: For our development, we utilised the `Humble' edition of ROS~2, a Long-Term Support (LTS) release, ensuring a stable foundation for our present and future experiments.
	\item \textbf{RoboStack}: RoboStack allowed for the integration of ROS with the Conda ecosystem, with a focus on package management. It helped ensure consistency across different systems during development when testing in simulation then on-car~\cite{FischerRAM2021}.
\end{enumerate}

\subsection{Simulated Environment}

The experiments utilised a simulator named EUFSsim, which is based on the Gazebo platform~\cite{EUFS_Simulator}. This simulator was originally developed by the Edinburgh University Formula Student (EUFS) team and later modified by QUT Motorsport. It is an extension of FSSIM for ROS~1. In addition to simulating vehicle dynamics, EUFSsim also emulates the sensors used on the vehicle (Section~\ref{sec:hardware_sensors}) and introduces simulated noise to enable realistic testing conditions.

The simulator was instrumental in generating quantitative results. It provided access to essential ground truth data such as maps, cone locations, and vehicle positions. This functionality contributed to the accuracy and reproducibility of the experiments, particularly considering the restricted access to the actual vehicle during the development phase.

\subsection{Real Life Track Experiment Setup}

Our real-life experiments were conducted in a controlled urban car park to simulate a segment of the target track specifications. The setting is depicted from the car's point of view in the bottom left of Figure~\ref{fig:custom_vs_ros_architecture}.

The car park environment differed from the FS standard track by the presence of background elements such as parked cars, trees, and poles. While not directly interacting with the vehicle, these elements were available for our camera's internal visual odometry during our tests. However, these elements will not be present in the actual competition, which could impact mapping and localisation accuracy.

In contrast to open tracks used in FS competitions, the urban car park was surrounded by tall buildings that likely affected the quality of GNSS data. Such architectural obstructions can interfere with satellite signals, leading to decreased GNSS accuracy. Given the absence of such obstructions on an open track, we anticipate that GNSS accuracy will be significantly better in a competition setting.

\subsection{Hardware and Sensors}
\label{sec:hardware_sensors}
The qualitative mapping and localisation evaluations were conducted on real hardware using the following sensors:

\begin{itemize}
    \item Velodyne Puck LiDAR, limited to 180 degree field of view,
    \item ZED2i Stereo Camera,
    \item SBG Ellipse-D INS,
    \item Wheel Speed Hall Effect Sensors, and
    \item Steering Angle Hall Effect Sensor.
\end{itemize}

\section{Results}

This section evaluates the chosen off-the-shelf ROS~2 navigation libraries in an applied way, determining suitability on our real vehicle and a quantitative way, comparing navigation metrics in simulation.

\subsection{Quantitative Metrics -- Simulation}

\begin{figure*}[!t]
    \centering
    \includegraphics[width=\linewidth]{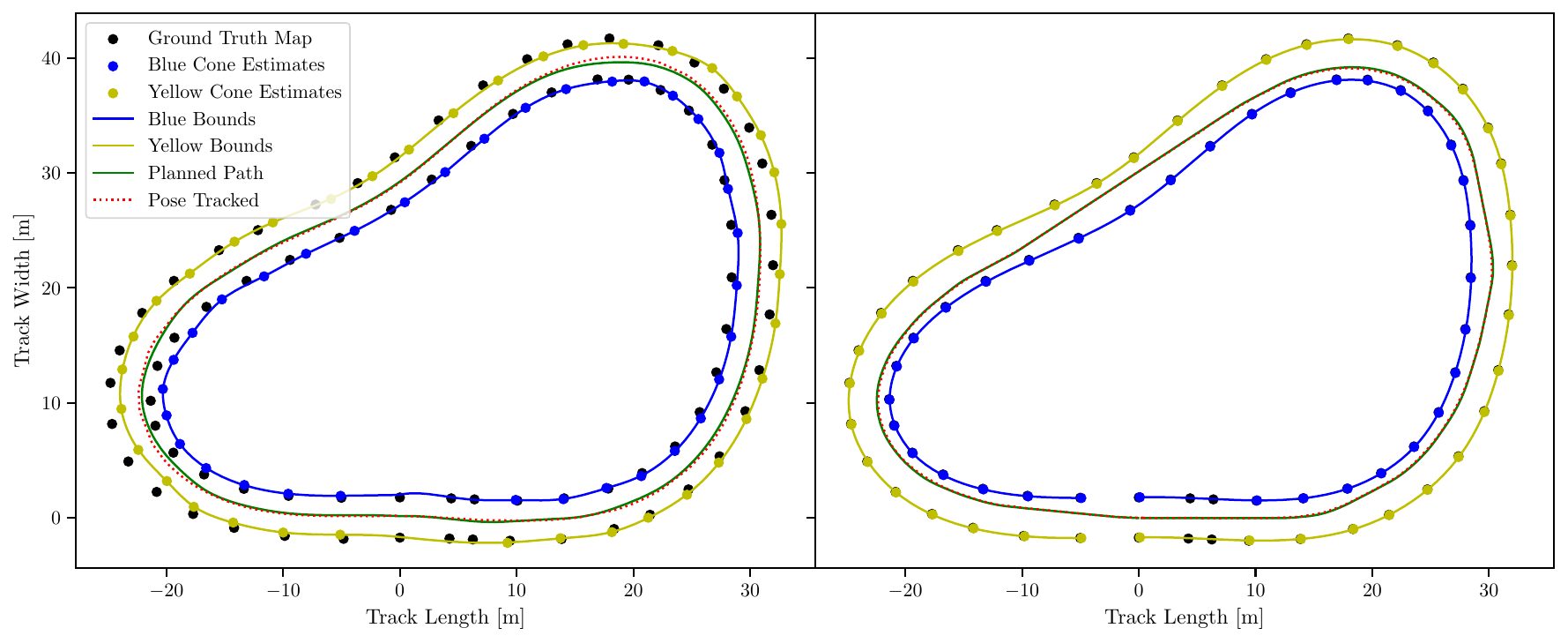}
    \caption{Navigation output of the previous custom stack developed by QUT Motorsport (left) and the new system utilising the ROS~2 navigation stack (right). This was the result of two laps: a discovery lap for mapping and a subsequent fast lap for guidance accuracy assessment. The previous approach showed substantial deviation from the ground truth, while the new approach yielded almost indistinguishable estimates from it.}
    \label{fig:double_image}
\end{figure*}

As outlined in Section~\ref{sec:methodology}, a variety of navigation metrics were compared between the newly integrated ROS~2 navigation packages and previous algorithms, with stages being evaluated layer-by-layer. These experiments validated our decision to move toward using off-the-shelf libraries and leverage superior performance with open-source software.

\subsubsection{State Estimation}

Here, our testing evaluates the effects of using \verb|robot_localization| for a strong state estimation on the output of our previous simple EKF landmark SLAM algorithm. We were looking to reduce the error of mapped cones and vehicle pose against ground truth data from the simulator. Results were recorded over four repeated individual laps of our simulated testing racetrack and can be seen in Table~\ref{tab:SLAM_results}.

Adding \verb|robot_localization| for a separate odometry filtering process before EKF landmark SLAM improved map accuracy by 65\% on average and pose tracking by 81\%. This increased accuracy would prove vital in ensuring cone landmark locations were consistent and less prone to incorrect detections, in turn, allowing for faster racing speeds.

The previous EKF landmark SLAM approach demonstrated its limitations in Figure~\ref{fig:double_image} with visible drift onwards from the first corner. Additionally, the algorithm had no internal loop-closure and struggled to re-identify cones in view on the map, particularly evident with cones missed as the vehicle approached its starting location. EKF landmark SLAM's susceptibility to incorrect detections was the primary factor which would lead to the re-evaluation replacement of this SLAM approach.

In the existing EKF landmark SLAM approach, our INS received GNSS updates at 5Hz, thus the predict step of the EKF was limited. With odometry filtering, the IMU and wheel encoders can operate at a much higher frequency and filtering was set to output at 50Hz in the configuration parameters. In turn, the vehicle position can be estimated more accurately over sharp movements that the GNSS prediction may miss. Having a separate odometry estimate updating at a much faster rate than a SLAM algorithm to refine the vehicle's position allows for faster control logic to react to changes in the track and desired racing line.
More importantly, this feature enables better utilisation of available data on QEV-3D by the navigation system, thereby significantly contributing to the reduction of RMSE.

\subsubsection{Mapping and Localisation}

Measuring the same metrics, our mapping and localisation testing evaluates how an advanced and robust SLAM algorithm within \verb|slam_toolbox| can further improve mapping and pose measures against the ground truth. This testing would compare \verb|slam_toolbox| to the previous EKF landmark SLAM with \verb|robot_localization| as outlined in our layered approach. Results, obtained over four laps, were also outlined in Table~\ref{tab:SLAM_results}.

The \verb|slam_toolbox| implementation proved far superior to its EKF landmark SLAM counterparts with an improvement of 79\% in mapping accuracy and 28\% in pose tracking accuracy over the previous stage. This meant using the \verb|slam_toolbox| over the team's previous SLAM approach equated to a 93\% mapping accuracy increase and an 87\% pose tracking accuracy increase, a far greater increase than we expected.

The sparseness of landmarks was the primary factor into the EKF algorithm’s limited ability to correct the motion prediction. In the test track, there were only 77 cones to designate the circuit boundaries and at any given time there would only be between five and ten to localise against. This was a limitation of the landmark SLAM that was in use and as a result, led to the conclusion being made about the limited effectiveness of the QUT Motorsport's previous solution.

Since the \verb|slam_toolbox| uses laser scans to match against known occupancy grid cells, it allows us to leverage longer scan range configurations than would be possible when detecting cones using landmark SLAM. Combining this with a fine grid resolution (5cm) means multiple occupied cells denote one cone and the solver can leverage a more distinctive map with more object points to reference against.

\subsubsection{Route Planning}

For route planning analysis, there was no ground truth to compare against, as the best path is subjective depending on the racing line configurations aimed at a particular vehicle. As outlined in Section~\ref{sec:method_planning}, the metrics we aimed to configure the \verb`smac_planner_hybrid` for was to minimise cumulative curvature in addition to minimising distance around the track. Routes would be planned after an initial discovery lap had identified the map and extracted boundaries. Results from this are shown in Table~\ref{tab:route_planning_comparison}.

On average, the \verb`smac_planner_hybrid` devised a lap that was 3\% shorter over the midline and reduced total curvature angle required to steer the vehicle by 5\%.
This would allow QEV-3D to maintain high speeds and not become unstable around corners in order to reach fastest lap times. Here, the reduced distance of 5m over this track whilst driving at a nominal speed of 8m/s would result in a predicted 0.62s faster lap time.

This stage required fine tuning of path penalties to provide the most optimal route while also maintaining a safe distance from bounds so any slight control deviations would not result in collisions with traffic cones.

\subsubsection{Guidance Control}
\begin{figure}[!t] 
\centering
\includegraphics[width=0.95\linewidth]{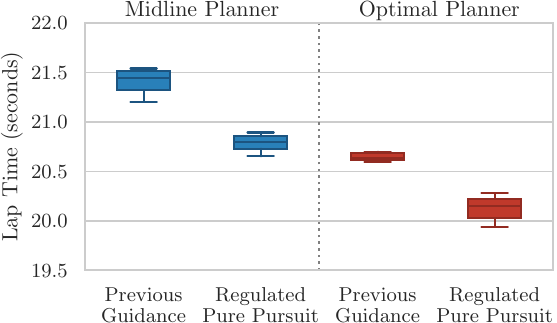}
\caption{Lap times across guidance and planning methods, showing a distinct decrease in overall lap-times as off-the-shelf ROS~2 packages are implemented.}
\label{fig:pure_pursuit_comparison}
\end{figure}

Similar to Route Planning, there is no `ground-truth' for us to compare with control, so, the new Nav2 guidance controller,  \verb`regulated_pure_pursuit`, was compared against the team's previous custom Pure Pursuit. Their relative performance were compared across both on our previous mid-line planner and the new  \verb`smac_planner_hybrid`, comparing cross-track following error and overall lap time. 
These results can be seen in Figure~\ref{fig:pure_pursuit_comparison}.
For consistency, the controllers target velocity and look ahead distances were kept the same at 8m/s and 2.5m respectively.

The \verb|regulated_pure_pursuit| ROS~2 package out-performed expectations, with a 0.7s average lap time improvement over the previous custom pure pursuit approach. This was surprising due to these two programs substantially similar nature. This 0.7s improvement was found to occur when racing on both route options. 

Additionally, these results validated the performance benefits of an optimal racing line generated by the \verb|smac_planner_hybrid|. Lap times did decrease by the predicted amount -- an average of 0.6s -- when using the racing line to reduce overall route distance and intensity of turns. Combining these Nav2 packages would improve QEV-3D's overall time by 1.5s, or 7\%.

\begin{table}[!t]
    \centering
    \caption{SLAM Comparison: Average cone detection and pose tracking error, showing improvements across accuracy metrics when replacing custom packages with off-the-shelf packages in the navigation system.}
    \label{tab:SLAM_results}
    \begin{tabular}{C{0.35\linewidth}C{0.27\linewidth}C{0.25\linewidth}}
        \toprule
        \textbf{SLAM Approach} & \textbf{Cone Detection RMSE [m]} & \textbf{Pose Tracking RMSE [m]} \\
        \midrule
        EKF landmark SLAM & $1.06 \pm 0.05$ & $0.46 \pm 0.03$ \\ \\
        EKF landmark SLAM + Robot Localization & $0.37 \pm 0.09$ & $0.09 \pm 0.02$ \\ \\
        SLAM Toolbox & $0.08 \pm 0.01$ & $0.06 \pm 0.01$ \\
        \bottomrule
    \end{tabular}
\end{table}

\begin{table}[!t]
    \centering
    \caption{Route Planning Comparison: Average distance and curvature, showing improvements across route efficiency metrics when replacing custom packages with off-the-shelf packages in the navigation system.}
    \label{tab:route_planning_comparison}
    \begin{tabular}{C{0.35\linewidth} C{0.27\linewidth} C{0.25\linewidth}}
        \toprule
        \textbf{Route Planning Approach} & \textbf{Total Distance [m]} & \textbf{Total Curvature [rad]} \\
        \midrule
        Custom midline & $ 149.96 \pm 0.01 $ & $ 13.841 \pm 0.715 $ \\ \\
        Smac Planner Hybrid & $ 145.95 \pm 0.48 $ & $ 12.722 \pm 0.098 $ \\
        \bottomrule
    \end{tabular}
\end{table}

When it came to cross-track RMSE, recorded in Table~\ref{tab:cross_track_error}, we see a slight decrease in error when \verb|regulated_pure_pursuit| is used over our previous approach, validating our decision to move away from the custom approach. This was demonstrated with an 8\% decrease when operating on the previous midline path and 9\% on the optimised path. 

Although the cross track error was higher for both controllers with the optimal planner over the midline planner, this was potentially due to the smaller resolution of path points being based on grid cells compared to the custom midline resolution based on a fixed points-per-cone ratio. Despite this slight increase, it is important to note that the vehicle did not deviate far enough from the desired line to cross the track boundaries at any time and collide traffic cones.

Pure pursuit algorithms will always have some element of corner cutting due to the nature of following a waypoint at a distance along the path. With the prior mid-line planning solution, this corner cutting could be leveraged as a way to actualise a more optimal line than what was originally planned. However, an optimal path combined with a fast-reacting controller could mean a shorter look-ahead distance, and thus, lower cross-track error as the vehicle more closely matches the path.

\begin{table}[!t]
    \centering
    \caption{Guidance Comparison: Average Cross-track RMSE, outlining the path following accuracy improvement when replacing custom packages with off-the-shelf packages.}
    \label{tab:cross_track_error}
    \begin{tabular}{C{0.35\linewidth} C{0.27\linewidth} C{0.25\linewidth}}
        \toprule
         \textbf{Guidance Approach} & \textbf{With Midline Planner} & \textbf{With Optimal  Planner} \\
        \midrule
        Previous Pure Pursuit & $ 0.101 \pm 0.002 $ & $ 0.122 \pm 0.002 $ \\ \\
        Regulated Pure Pursuit & $ 0.093 \pm 0.002 $ & $ 0.112 \pm 0.016 $ \\
        \bottomrule
    \end{tabular}
\end{table}

\begin{figure}[!t] 
\centering
\includegraphics[width=0.9\linewidth]{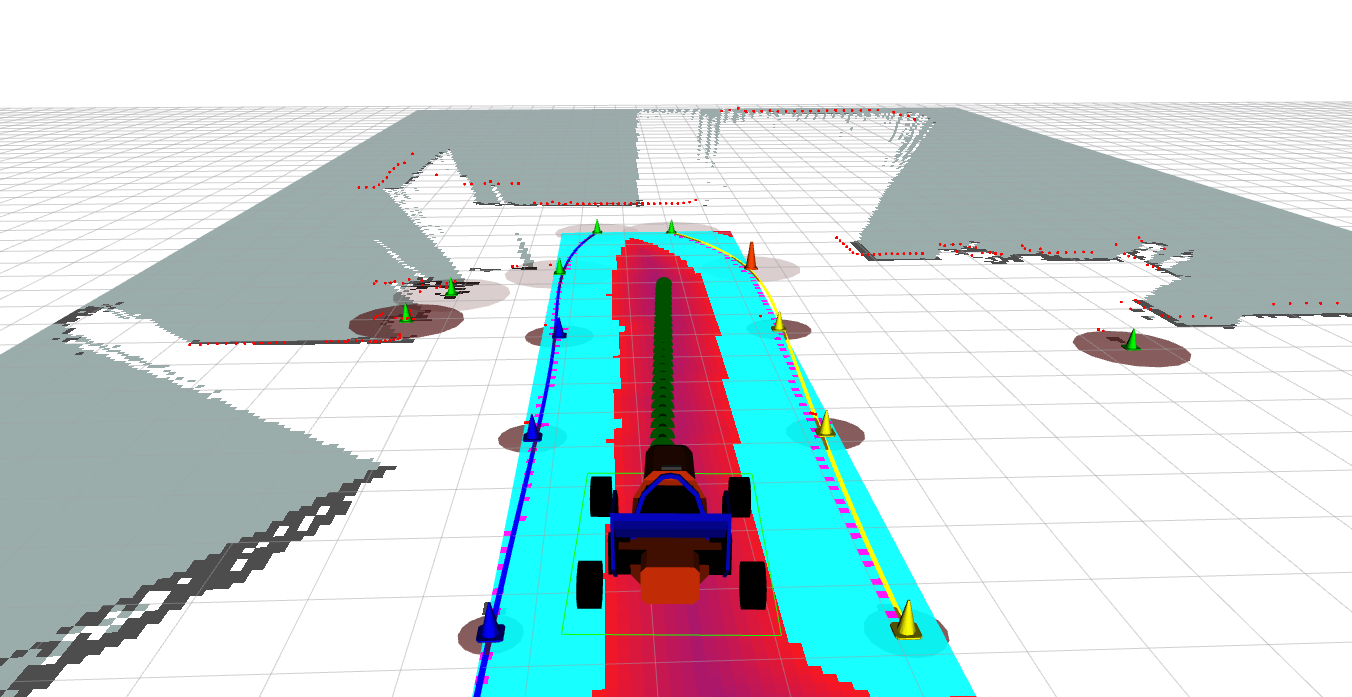}
\caption{ROS~2 Navigation stack operating on QEV-3D in a carpark test, as seen in Figure~\ref{fig:custom_vs_ros_architecture} from its point-of-view.}
\label{fig:carpark_nav}
\end{figure}

\subsection{Applied -- On vehicle}

The ultimate aim of this project was to demonstrate the capabilities of ROS~2 and Nav2 for off-the-shelf software support for a real autonomous race car. This involved installing the ROS~2 navigation stack with our custom interfaces on QEV-3D for some limited running in a real environment at the time of writing.

Our testing took place in a small carpark, so the scope of complexity was limited, but we were still able to run the whole stack from end-to-end, utilising all available sensors and replacing all existing navigation components. As part of this, guidance control was running, but given our testing environment was in a carpark, we were unable to examine the driving output of the navigation loop. This was visualised in Figure~\ref{fig:carpark_nav}.

\subsubsection{Sim-to-real barriers}

Unlike the Gazebo simulator, QEV-3D was limited in ROS~2 driver support for some of its available sensors, namely, wheel and steering angle hall effect sensors. These devices were read by a micro-controller on our vehicle control unit and information sent to our ROS-running computer. This analogue-to-digital voltage data needed to be converted into ROS messages which could be used by the \verb|robot_localization| filter as these important odometry measuring sensors would drastically improve the accuracy of the filtered state estimate. Thus, a custom hardware interface was developed to run on-board QEV-3D only and was not required when running the stack in our simulator.

Additionally, our ZED2i camera's internal odometry tracking proved limited in forward motion -- likely due to features appearing similar from frame-to-frame. As a result, it was decided that only angular motion would be utilised in the \verb|robot_localization| filter, which proved reliable. We updated our simulated stack to replicate this.

Despite limited testing and some minimal changes to bridge the sim-to-real gap, it is apparent that this stack will outperform the previous navigation system on on the real vehicle. A full testing regiment for validation is only a matter of taking QEV-3D out to a racetrack.

\section{Conclusions and Future Work}

This paper offered a detailed tutorial on integrating a ROS~2-based navigation stack into a Formula Student autonomous vehicle, with special emphasis on QUT Motorsport's QEV-3D. Our methodology was delineated into four pivotal phases: state estimation, mapping and localisation, route planning, and guidance control. Each phase aimed to either replace or augment existing custom components of the vehicle’s navigation system. The objective was to realise incremental but significant improvements in key performance metrics such as mapping accuracy, route efficiency, and path-following precision.

The adoption of off-the-shelf packages, most notably from the Nav2 library, underscored the merits of leveraging mature, open-source software. This approach not only accelerates the development cycle but also enhances the robustness and performance of the system.

For future work, one avenue to explore is the investigation of alternative SLAM techniques. This could include delving into 3D SLAM or even hybrid models that amalgamate both 2D and 3D mapping techniques. Another area of interest is the evaluation of different planning algorithms, as the existing options have not fully met our specific requirements. Further improvement can be achieved by modifying the occupancy grid generated by the \verb|slam_toolbox| to accommodate objects detected by laser scans in addition to cones. This approach would obviate the need for creating an entirely new grid centred solely on cones. On the control side, different controllers like model predictive control could be assessed, especially those that take vehicle motion modelling parameters into account. Lastly, forking, modifying, and contributing back to the open source repositories that were used for this work could be a strategic move to reduce the dependency on custom converters and other bespoke elements, thereby streamlining the development process.

\bibliographystyle{IEEEtran}
\bibliography{references}

\end{document}